# A Clustering Analysis of Tweet Length and its Relation to Sentiment

Matthew Mayo
TSYS School of Computer Science, Columbus State University
mayo_matthew@columbusstate.edu

Abstract

Sentiment analysis of Twitter data is performed. The researcher has made the following contributions via this paper: (1) an innovative method for deriving sentiment score dictionaries using an existing sentiment dictionary as seed words is explored, and (2) an analysis of clustered tweet sentiment scores based on tweet length is performed.

Keywords: sentiment analysis, Twitter, Weka, clustering, machine learning, data mining, natural language processing

## 1. Introduction

Twitter[1] is a popular microblogging web service that allows users to share and view short messages online, and which also exhibits a number of social networking characteristics. Started in 2006, Twitter today has over 645 million active registered users who generate on average 58 million messages, or tweets, per day [9]. With over 9,000 tweets being dispatched every second from all around the globe, the Twitter data stream has the potential to provide previously unimaginable insight into individuals' views on a wide range of topics.

Sentiment analysis refers to the use of technology for the identification, extraction and processing of subjective information from source material, including attitudes, emotions and opinions [6]. Often referred to as opinion mining, sentiment analysis makes use of natural language processing, machine learning and statistical analysis in its processing of source materials. Source materials can be any human language construct ranging from product reviews to blogger election attitudes [6]. Numerous strategies exist for analyzing the sentiment of subjective text; however, applying binary classification to terms (good vs. bad, positive vs. negative) and using a scale-based system to assign scores to terms are 2 popular approaches.

The allowable tweet limit of 140 characters imposed by Twitter provides a unique venue for users to convey their succinct ideas and opinions to world. The enormous popularity and extensive data streams of Twitter provide a massive opportunity from which to draw upon for analysis. There is little wonder why Twitter has become a popular target for sentiment analysis. In this paper, the researcher aims to explore the relationship between the length of a tweet in number of characters and its sentiment score.

---

[1] https://www.twitter.com

After capturing and processing tweets, length and sentiment score pairs will be clustered using a machine learning algorithm in order to explore whether relationships exist in the data. The goal is to determine whether the number of characters a tweet is composed of relates to a tweet's sentiment score.

## 2. Research Description

A large body of innovative Twitter sentiment analysis exists, including research in areas such as political elections [1], product reviews [2] and the fluctuating mood of Twitter users across time [5]. However, there seems to be very little research into the area that this paper will focus on.

As a first step in this research project, the Twitter Tweet Application Programming Interface (API)[2] is used to tap into the Twitter stream and collect user tweets in real time for later processing. Next, a well known sentiment analysis dictionary[3] is used as a starting point to creating a custom sentiment dictionary based on the content of our captured tweets. Tweet lengths are then computed and are assigned a sentiment score based on the sum of the individual term scores as per the sentiment dictionary. Then the dataset to be used for clustering is built,
which includes the length of a tweet, its computed sentiment score and the date of capture. Finally, the dataset will be loaded into a well-known machine learning workbench, Weka[4], to be clustered using the k-means algorithm. It is hoped that this process will provide insight into whether or not tweet length and tweet sentiment are related.

Though a large amount of data was captured from the Twitter stream during the capture sessions, the only tweets considered for this research project were those written in English and originating in the United States as per the city attribute of the Tweet API place field. The inclusion of tweets beyond these criteria would lead to complications and incomparable results.

## 3. Data Description & Processing

In order to capture Twitter tweets, one must first register an application[5] with Twitter in order to receive the information that is required for authenticating to Twitter via their API. Once an application was registered and authentication information received, the researcher used this data in a custom Python script[6] written to capture Twitter tweets from the live stream. Tweet data was subsequently stored in the JavaScript Object Notation, or JSON, format and saved to file. Tweets were captured during 4 one-hour sessions per week, over the course of 3 weeks. The times of the week selected for capturing the data were based on previous research investigating mood throughout the day inferred from Twitter [5]. Times around the 2 happiest points of the week as well as the 2 least happy points of the week were selected for data capture.

---

[2] https://dev.twitter.com/docs/platform-objects/tweets
[3] http://www2.imm.dtu.dk/pubdb/views/publication_details.php?id=6010
[4] http://www.cs.waikato.ac.nz/ml/weka
[5] https://apps.twitter.com
[6] https://github.com/mmmayo13/tweet-sentiment-scores

This strategy was employed to ensure sentiment representation from potentially both happier and less-happy tweets. Each data capture output session resulted in a single large file of several hundred megabytes, and stored a Twitter specified maximum number of tweets per time period that a user is able to extract from the live stream. In order to assign sentiment scores to a particular corpus of text, a sentiment dictionary composed of individual terms and their sentiment scores is required to compare the corpus terms to. An existing, well known dictionary is AFINN-111 [7]. It provided a well-tested starting point for building a custom sentiment dictionary.

Domain specific sentiment lexicons have been shown to be more accurate than general dictionaries [4]. With inspiration from Rice & Zorn [8], the researcher built a custom corpus sentiment dictionary using AFINN-111 as a set of sentiment score seeds. The researcher wrote a custom Python script which took as input a predefined sentiment dictionary and iterated through a file of captured tweets, computing sentiment scores for terms not included in the existing dictionary. All captured tweets were merged into a single file for input. A simple, mean-based formula was used to compute new sentiment term scores, such that the new term sentiment score for any non-scored term in a particular tweet is calculated as the sum of sentiment scores of scored terms in this tweet divided by the number of terms in this tweet.

The underlying assumption was made that non-scored terms in a tweet would be related in sentiment to previously scored terms. Assigning non-scored terms the mean of the sentiment scores of all other terms in the tweet was the relation chosen for implementation.

The output of this process was then merged with the original sentiment dictionary, AFINN-111, and the results became the custom corpus sentiment lexicon for the captured tweet data. Once captured data was available and a custom dictionary had been built, tweets were to be processed one by one to obtain their sentiment score and their length in number of characters. A custom Python script was created by the researcher to iterate through a tweet data file and assign each tweet a sentiment score based on the sum of the scores of its individual term sentiment scores. This sentiment score, along with the tweet length in characters, were output to a comma separated value (CSV) file.

The processed output files were then concatenated into a single file, and the samples randomized. As this dataset was to be read by the Weka machine learning workbench, the randomized CSV file was then converted to an Attribute-Relation File Format (ARFF)[7] file by adding the appropriate header information. The resulting dataset contains 14763 instances of tweet lengths and corresponding sentiment scores. Using Weka, the researcher loaded the dataset and clustered the instances via the k-means clustering algorithm. k-means is a centroid-based clustering technique that partitions a dataset into clusters by assigning an instance to the cluster with the nearest mean [3]. As clustering is an unsupervised machine learning technique, once the dataset is loaded and the algorithm executed, the results are ready for inspection.

---

[7] http://weka.wikispaces.com/ARFF

## 4. Experiments & Discussion

The researcher trained a simple k-means clustering model in Weka. By accepting all default values, which included building 2 clusters using all attributes, the dataset was split 67/33 percentage-wise, and 9950/4813 by number of instances. In Table 1, below, cluster 0 corresponds to shorter tweets, with a centroid of approximately 41 characters long, which have a centroid sentiment score of approximately 1.2. Cluster 1 corresponds to longer tweets, with a centroid of approximately 104 characters long, which have a sentiment score centroid of approximately 3.8.

| Attribute | Full Data | Cluster #0 | Cluster #1 |
|---|---|---|---|
| length | 61.7673 | 41.1478 | 104.3943 |
| sentiment | 2.0732 | 1.2322 | 3.8120 |

Table 1: Centroid Summary by Cluster

As is visible by the cone shape of Figure 1, it appears that sentiment scores are initially tightly centered on the cluster's centroid, and as tweet length increases, the probability of any particular tweet being assigned a sentiment score further away from the cluster's centroid, in either direction, becomes greater. Sentiment scores of shorter tweets appear more tightly-centered around their cluster's centroid, while longer tweets become less-centered on the applicable centroid. As the number of characters in a tweet would lead to a greater number of terms, which would increase the chances of terms being assigned a score, this seems intuitive.

Let us consider that there is no relationship between tweet length and sentiment. We would expect that as the length of a tweet grew longer, the occurrences of positive and negative terms would be relatively equal in probability, and would cancel one another out. If there were no relationship between length and sentiment, this should hold true regardless of tweet length. However, since tweets display greater variability in their sentiment scores as they increase in length, it would seem that there is a relationship between tweet length and tweet sentiment score.

This relationship does not affect the predictability of whether a sentiment score will be positive or negative; however, it does affect the probability of how far the sentiment score will be from a particular centroid. This is, again, intuitive, since a single, focused, coherent thought from the human mind will have a tendency to progress linearly in its reasoning. Any progressive reasoning is likely to carry the previous portion of its argument forward. Hence, a positive thought, idea or opinion is likely to continue on a continuous trajectory to its completion. The same would be true for a negative thought, idea or opinion. Likewise, a tweet beginning with a positive sentiment is likely to continue on that course for as many characters as it exists for, and conversely for a negative tweet. While positivity begets positivity, the opposite is also true. Increased tweet length provides for the additional opportunity for such like sentiment to continue, which should send the accumulating sentiment score further in the same direction, and away from a cluster's centroid, or from the cluster's mean. This is a generalization, of course, one that is both intuitive and nearly impossible to prove in any quantitative manner.

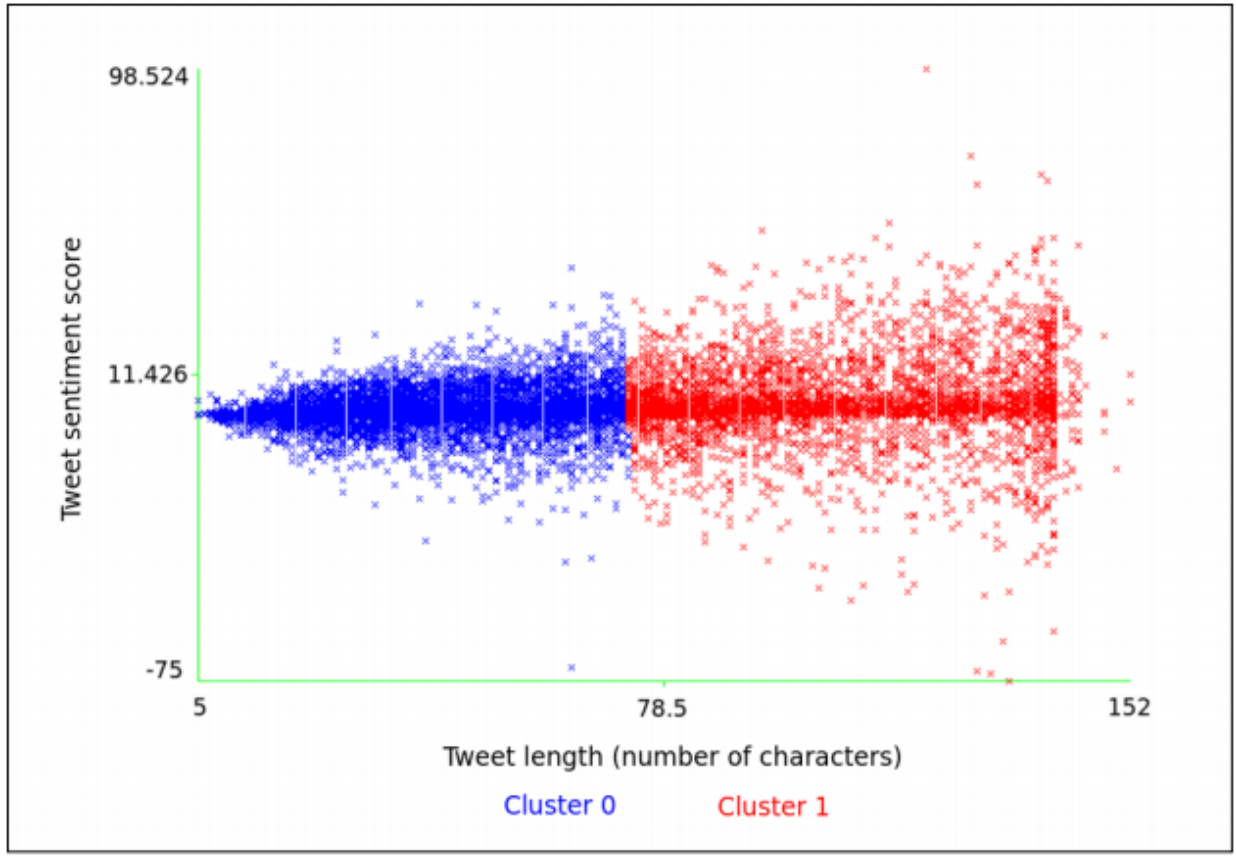

Figure 1: Clustered Tweet Sentiment Scores

While relatively little treatment has been given to the particular implementation decision details, it should be noted that crafting a custom sentiment dictionary is an area of research where no consensus has been reached as to which situation calls for what solution. It is an expanding field of research, and there are numerous problems associated with building a custom sentiment dictionary of any type.

These findings should serve as notice that any practical sentiment analysis algorithm generalizations should be taking tweet length into account. For example, any political election result predictions based on Twitter sentiment analysis should consider correcting for tweet lengths. If tweets of favorable sentiment are of significantly different mean length than tweets of unfavorable sentiment, bias may be an issue. Further research could focus on expanded Twitter data, as well as alternate in-depth strategies for creating custom sentiment dictionaries.

# 5. References

[1] Bakliwal, A., Foster, J., van der Puil, J., O'Brien, R., Tounsi, L. & Hughes, M. (2013). Sentiment analysis of political tweets: Towards an accurate classifier. *Proceedings of the Workshop on Language in Social Media 2013*, 49-58. http://www.mpisws.org/~cristian/LASM_2013_files/LASM/pdf/LASM06.pdf

[2] Bukherjee, S. & Bhattacharyya, P. *Feature specific sentiment analysis for product reviews*. Retrieved from http://www.cse.iitb.ac.in/~pb/papers/cicling12-feature-specific-sa.pdf

[3] Han, J., Kamber, M., & Pei, J. (2012). *Data mining: Concepts and techniques* (3rd ed.). Waltham, MA: Morgan Kaufmann Publishers.

[4] Minocha, A. & Signh, N. (2012). Generating domain specific sentiment lexicons using the Web Directory. *Advanced computing: An international journal*, 3(5), 44-51. http://airccse.org/journal/acij/papers/3512acij05.pdf

[5] Mislove, I., Lehman, S., Ahn, Y., Onnela, J. & Rosenquist, N. (2010). *Pulse of the nation: US mood throughout the day inferred from Twitter*. Retrieved from http://www.ccs.neu.edu/home/amislove/twittermood/

[6] Mullen, T. (2012). *Introduction to sentiment analysis* [PDF document]. Retrieved from http://www.lctmaster.org/files/MullenSentimentCourseSlides.pdf

[7] Nielsen, F. A. (2011). A new ANEW: Evaluation of a word list for sentiment analysis in microblogs. *Proceedings of the ESWC2011 Workshop on 'Making Sense of Microposts': Big things come in small packages 718 in CEUR Workshop Proceedings*, 93-98. http://arvix.org/abs/1103.2903

[8] Rice, D. R. & Zorn, C. (2013). *Corpus based dictionaries for sentiment analysis of specialized vocabularies*. http://www.kenbenoit.net/pdfs/NDATAD2013/Rice-Zorn-LSE-V01.pdf

[9] Statistics Brain. (2014). *Twitter statistics*. Retrieved from http://www.statisticbrain.com/twitter